# Rapid Training Data Creation by Synthesizing Medical Images for Classification and Localization

Abhishek Kushwaha     Sarthak Gupta     Anish Bhanushali     Tathagato Rai Dastidar
{abhishek.k,sarthak,anish,trd}@sigtuple.com

## Abstract

*While the use of artificial intelligence (AI) for medical image analysis is gaining wide acceptance, the expertise, time and cost required to generate annotated data in the medical field are significantly high, due to limited availability of both data and expert annotation. Strongly supervised object localization models require data that is exhaustively annotated, meaning all objects of interest in an image are identified. This is difficult to achieve and verify for medical images. We present a method for the transformation of real data to train any Deep Neural Network to solve the above problems. We show the efficacy of this approach on both a weakly supervised localization model and a strongly supervised localization model. For the weakly supervised model, we show that the localization accuracy increases significantly using the generated data. For the strongly supervised model, this approach overcomes the need for exhaustive annotation on real images. In the latter model, we show that the accuracy, when trained with generated images, closely parallels the accuracy when trained with exhaustively annotated real images. The results are demonstrated on images of human urine samples obtained using microscopy.*

## 1. Introduction

Deep learning applications in the field of Medical Image domain is marred by the problems of accurate data generation since it requires expert knowledge. Urine Microscopic images are dense, and annotating every object with high accuracy is a time-consuming and costly affair. In life sciences, only certified experts are allowed to annotate data for any product development. Generating accurate annotated data to the order of tens of thousands of images in this domain poses a significant challenge to the fast development of models.

Approaches like the use of graphic simulators to generate automatically labeled data have been developed in other domains like indoor scene classification [6] etc which have shown good results but there is a lack of simulators for medical fields. Also, simulators have been successful for big objects but it is not clear how well they will perform with small objects as small objects have very few discriminative features.

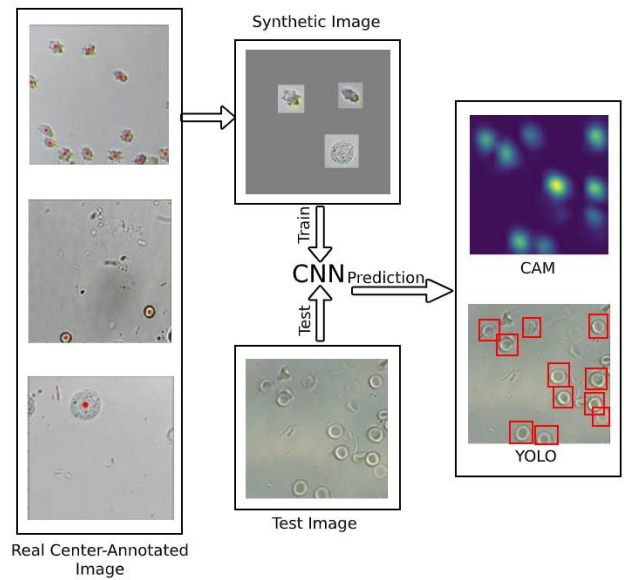

Figure 1. Proposed training methodology for training a deep neural network enables better feature learning while reducing the labour intensive work of annotation for strongly supervised learning. Center-annotated data by experts (left) is used to create synthetic data (middle-top) and corresponding labels. We use CAM (right-top) to perform localization. We also directly train on this synthetic data for object detection task (right-bottom).

Another way to tackle the problem of availability of correctly labelled data is by using weakly supervised training (multi-label classification) as they require annotations that can be generated with speed, and after training, the counting of objects is achieved using class activation maps (CAM) [19]. We show that even when the F1-score of these models for object classification is very high and numerically close to the proposed method, CAM generated heatmaps are not aligned with accurate object localization.

Here we present a simple process for quickly generating synthetic images to train a deep neural network for classification and object detection as illustrated in Fig. 1. The main contribution of this paper extends to the following. 1. Devising a synthetic data generation method without the need for any real background images. 2. Creating a synthetic image with such an object distribution which enhances feature learning. 3.Practical applications of such a technique in the medical domain for urine analysis.

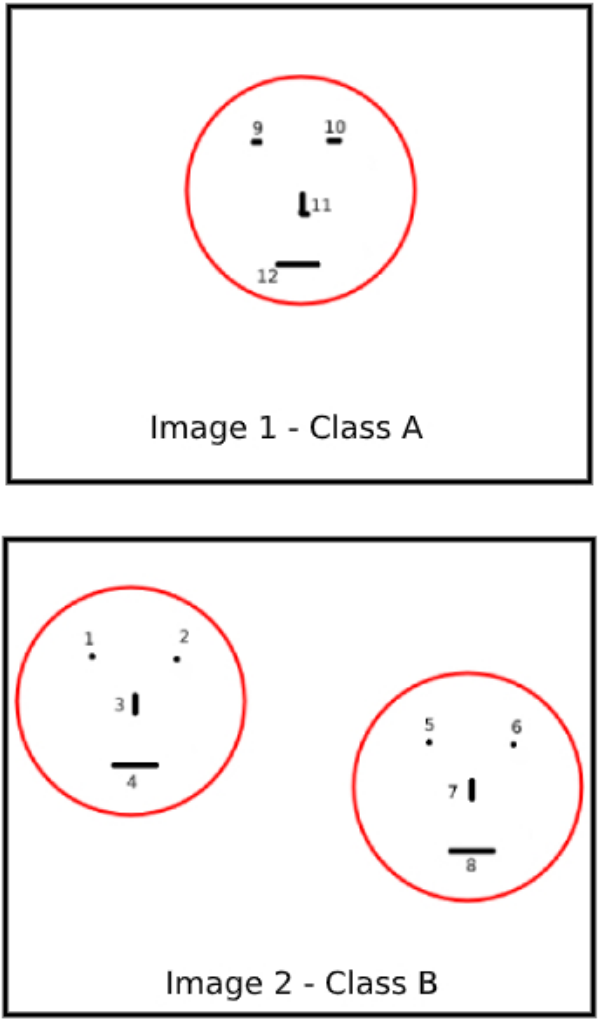

Figure 2. Toy example explaining why discriminative feature learning can not guarantee object level feature learning.

As shown in section 4, our method produces competitive results when trained on a practical task. For example, our method produces visually correct and explainable CAM [19] as compared to a model trained only on real images. Furthermore, for object detection, on mixing 10 percent of real data with synthetic data, our results show that in several cases we achieve better results than the models trained only on costly annotated real data

## 2. Related Work

Training deep neural networks using synthetic training data has gained popularity in the recent past and there exist many synthetic datasets like Flying Chairs [3], MPI Sintel [1], UnrealStereo [18], SceneNet [6], SYNTHIA [14], Sim4CV [11], and Virtual KITTI [4] among others. Every dataset is built for a specific task and varies in the process of creation.

We propose a method to synthesize Training images in such a way as to artificially change the distribution of objects in the images producing CAMs, for weakly supervised learning, which is correctly aligned with object localization. At the same time, this method overcomes the need for exhaustive annotations for a strongly supervised algorithm contributing to the speed up of model development.

The work of Dwibedi *et al.* [2] uses synthetic data to train for indoor object detection. Our work is conceptually closest to this work which involves extracting objects of interest from original images and using a deep learning model to blend it with another real background image, creating new images. Our approach is much simpler and different from this work. We use a simple extraction method to extract the object and stitch it onto a new image. Also, most of the synthetic data creation work involves using a real background which adds to the complexity of synthetic data creation but our method does not require any real background image thus further simplifying the process.

The use of synthetic data for extracting text from images has been explored by Gupta *et al.* [5] in which they overlay synthetic text to existing background images in a natural way, accounting for the local 3D scene geometry, to train the model.

Similar approaches have been explored to train robotics control policies [8, 17, 9]. Training object classifiers from 3D CAD models have been explored by [12]. The advantage of such methods is that one gets an unlimited amount of labeled data in a short duration of time but creating such a simulator to produce data is costly and difficult as the artists need to model every environment in detail.

## 3. Synthetic-data

We begin by explaining why the model may fail to learn individual object level features and instead learn group features. Then we introduce our image recreation process which overcomes this problem and forces the model to focus on object-level features. Then we show how this method overcomes the need for exhaustive annotation for object detection model training, speeding up the product development cycle.

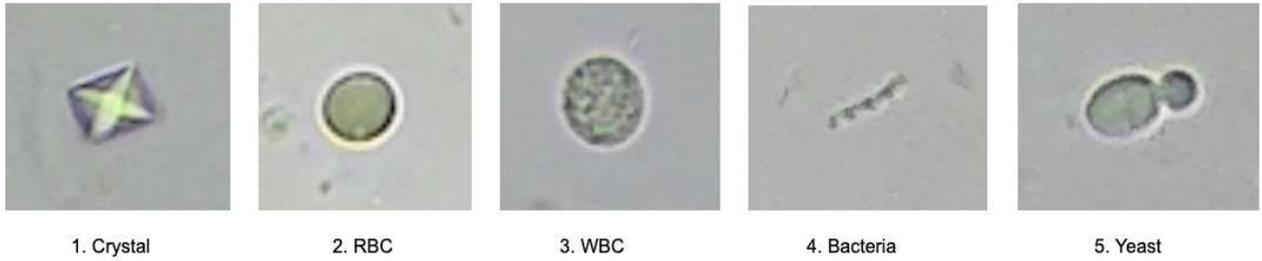

Figure 3. Sample objects which are important for Urine-analysis extracted from FOV images (Fig. 4,5). As can be seen, the bacteria class consists of very small objects and has very less visual features leading to even less discriminative features which makes learning difficult

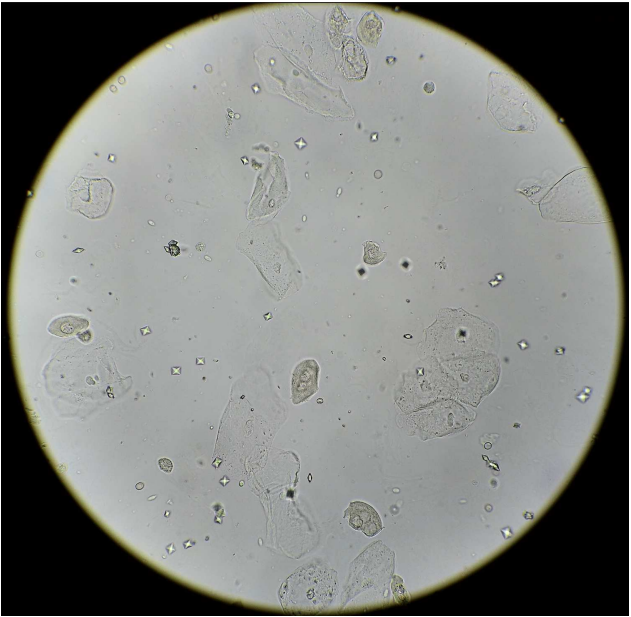

Figure 4. A microscopic field of view (FOV) of a urine sample. It can be observed that there are many objects which are artifacts without any fixed geometry and model needs to learn to discriminate it from objects of interest.

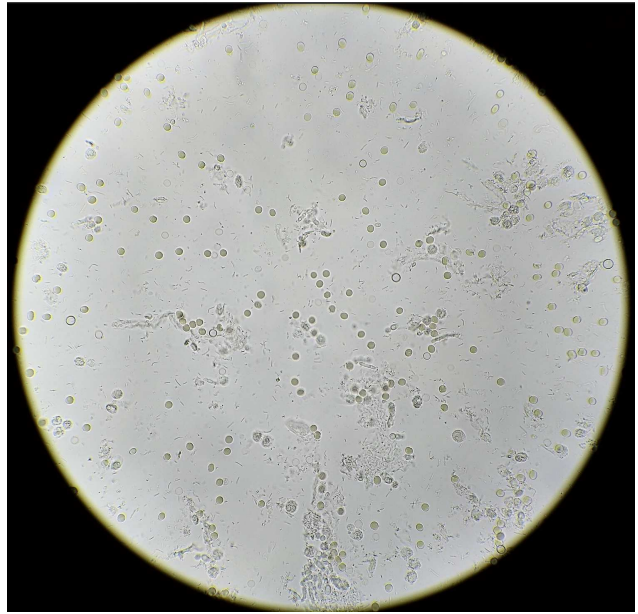

Figure 5. A microscopic field of view (FOV) of a dense urine sample with the presence of many objects of interest. Notice that bacteria is difficult to discriminate from artifacts even with human eye. Also, exhaustive annotation of objects in such images is time consuming and prone to excessive human errors.

### 3.1. Object and discriminative features

Deep learning models learn through gradient descent with cross-entropy loss. This loss forces the model to learn discriminative features. Zeiler *et al.* [16] have shown that initial layers learn low-level features while deeper layer learns high-level features. However assuming that high-level features will be object-level features may not be correct. This is due to the fact that the model has no concept of *object* and learns only *discriminative* features from the training set images. Due to a large number of images and variations in the training set, gradient descent will make the model converge to object-level features but such convergence is not guaranteed. Additionally, when the number of images are less and the number of objects high in each image, this convergence may become distant, as there can be more features that are discriminative but not necessarily individual object features.

To understand this consider a simple example with just 2 images in the training dataset as shown in Fig. 2. In image 1, we have a class A object and we can say its mid-level features are eye (9,10), nose (11), and mouth (12). In image 2, we have two class B objects. Their features are also numbered. If a model is trained using these two images, then we may assume that the model will learn features that are discriminative which in this case are eyes and nose. But the model training objective is to discriminate between two images and not two objects, so it may learn features (1,2,5,6) combined as a single feature for image 2 and features (9,10) for image 1 to differentiate two images, thus producing a low loss value. We have no control over forcing the model to learn individual-level features. If we use more images of both classes in different settings, the model will eventually be forced to learn individual-level features, but that objective is dependant on the distribution of individual objects in images, and not guaranteed in general. (This is one of the reasons why deep learning models need a lot of data to generalize better). The distribution of objects in the image happens to influence the *object-level* feature learning and since one cannot control the object distribution in real images, weakly supervised training with CAM localization does not produce acceptable and explainable results as shown in Fig. 8 (left column).

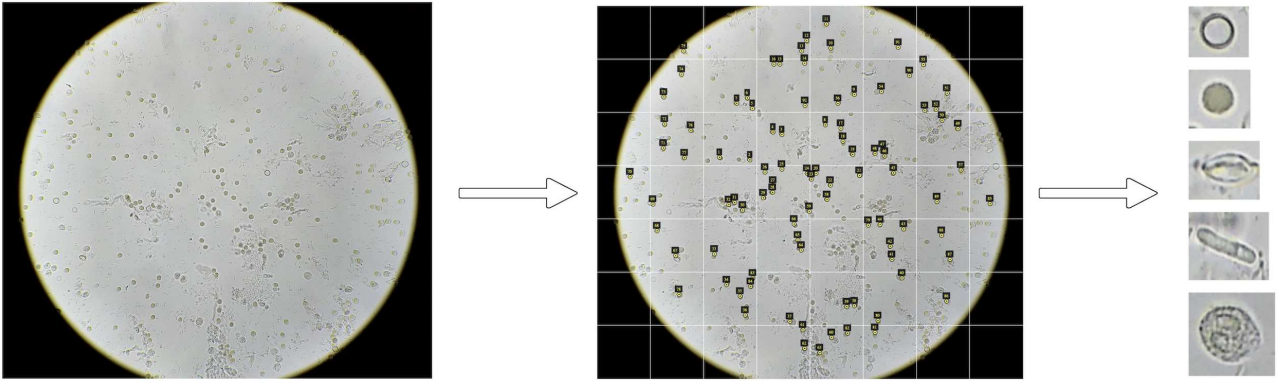

Figure 6. FOVs are non-exhaustively centre-annotated and labelled by medical experts. A patch is cropped around the annotation and extracted. This gives us a list of patches of objects which will be further stitched on the blank images to create synthetic images.

## 3.2. Data and Annotation

Sample urine microscopic images used are as shown in Fig. 4,5. For each urine sample, we take around 30 field-of-view (FOV) images by digital microscope. For our two experiments (weakly supervised and object detection) we take a random subset from 290 different urine samples. We focus only on the following medically significant urine objects- Bacteria, Crystal, Red Blood Cells (RBC), White Blood Cells (WBC), and Yeast as shown in Fig. 3.

### 3.2.1 Data for weakly supervised algorithm

Each object in the FOV was marked using the center annotation technique (marking only the center of the object and labeling them instead of making a complete bounding box around the object) by medical experts without focusing on the exhaustiveness of annotations in the image. Synthetic images were then created as described in the section 3.3. For weakly supervised model training, around 10k synthetic images were created. For comparison with model trained on real images, a single FOV (4072 X 3072 pixels) was sequentially cropped into many patches with each patch size equal to 384X384 pixels. Each patch was multi-labelled by medical experts with respect to presence or absence of each class.

### 3.2.2 Data for object detection algorithm

Around 7.7k synthetic images of size 416X416 pixels were created for training a model on synthetic data. Similarly, for comparison with a model trained on real images, around 7.7k real images of size 416X416 pixels were cropped from the original FOV dataset and exhaustively annotated by medical experts (this took 10X more time as compared to center annotation).

## 3.3. Synthetic-Data creation Process

To force the model to focus on *object-level* feature learning, we automatically generate synthetic images for model training from original images using the following process:-

1. From the original microscopic image, a patch is cropped using the marked center of object. The size of the patch is equal to the average biological size of that object. The objective is to extract a patch that contains only that particular object.

2. A blank image of the required size (either 384X384 or 416X416 depending on the training task) is created.

3. Extracted objects from step 1 are selected for creating a new image. The number of total objects, and the number of objects of each class which will be pasted are sampled with equal probability from the training dataset.

4. For this new synthetic patch, labels of pasted objects and their bounding boxes are generated from the original annotations done by medical experts. For weakly supervised training, a label for the whole synthetic image is created, while for object detection, we use each

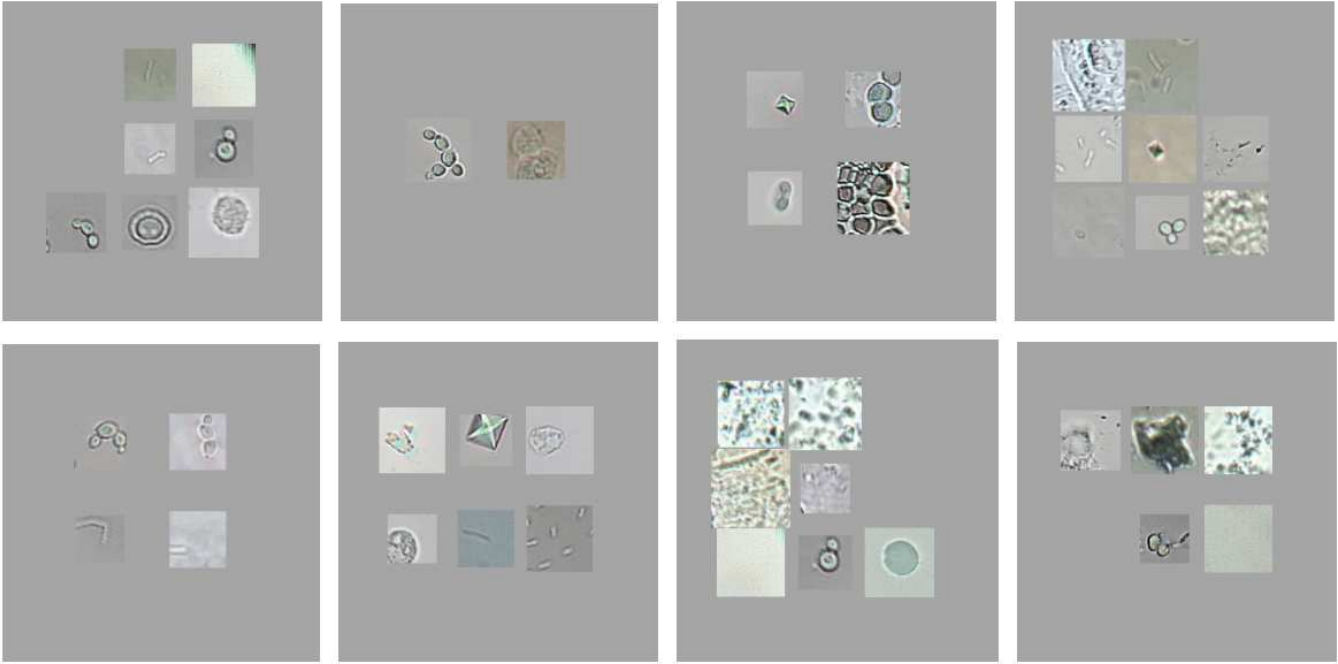

Figure 7. Samples of generates synthetic images. Number of patches in an image are random. Similarly classes present in an image are also random. During training, these images are generated on the fly.

   pasted object's bounding box and its respective marked label.

5. Before/after pasting patch on blank image, all relevant kinds of standard image augmentations are applied on extracted-object/new-image.

Fig. 6 encapsulates steps 1-4 of the data creation process. Fig. 7 displays some of the synthetically created images. Note that the distribution of the number of objects, and the number of classes of objects is random. Also, for the model to discriminate between objects of interest and universe, we have included patches that do not contain any object of interest but just artifacts. Medical experts annotated artefacts as a separate class while labeling objects. Artefact patches were extracted in the same way as objects of interest patches. Hence, six classes of objects were annotated and extracted instead of the five medically significant classes of urine objects.

It is to be noticed that at times whole of the object is not extracted as the extraction size is numerically equal to the biological average size of that particular class of object, with a 0-10% variation in size, which is randomly chosen. This is done to avoid any bias and unwanted pattern creation while creating the synthetic data. At the same time it also helps in regularization. Its regularization property can be understood in-terms of dropout. In this process 0-10% of the object is randomly dropped and the model is forced to learn about the object from its parts rather than from the whole object, which in turn helps in generalization.

| Objects | Train Real | Train Synth. | Test Real |
|---|---|---|---|
| RBC | 6178 | 5428 | 623 |
| YEAST | 4055 | 3064 | 233 |
| CRYSTAL | 3050 | 2710 | 296 |
| WBC | 4728 | 4746 | 938 |
| BACTERIA | 3852 | 3871 | 2007 |

Table 1. Distribution of class objects in real, synthetic and test images. Train and Test split is done on urine-sample level to ensure no overlap of data distribution between training and testing.

## 4. Evaluation

To evaluate the performance of our method, for weakly supervised training, we compared CAM generated using the model trained with the proposed method to CAM generated from the model trained on real images. In both cases, the test dataset was kept the same but different from training. For object detection algorithm, precision and recall values of predicted bounding boxes were compared between model trained on synthetic images and model trained on real images.

### 4.1. Weakly supervised algorithm

We used a network which has VGG16 [15] as the backbone architecture with some additional branches (experimental observation showed that this network produced better CAM [19] as compared to VGG16 [15], Densenet [7] etc). We trained it as a multi-class, multi-label model for 5 classes with binary labels for each class using the sigmoid function in the last dense layer of our network. We generated a class activation map for localization as described by Zhou *et al.* [19]. Fig. 8 shows the CAM generated by two models for different objects on the test set. Also, Table 2 shows the F1-score for the classification problem. Note that the model predicts only the presence or absence of an object in an image irrespective of it's count. F1-score is calculated only on this metric.

| Object | Real | Synthetic |
|---|---|---|
| RBC | 0.75 | 0.72 |
| YEAST | 0.61 | 0.65 |
| CRYSTAL | 0.79 | 0.79 |
| WBC | 0.85 | 0.83 |
| BACTERIA | 0.71 | 0.76 |

Table 2. F1-scores for weakly supervised training (multi-label, multi-class). These are classification scores.

High F1-score of RBC (.75) for real images model shows that the model has learned to discriminate it from other classes in the image. Hence we expect the model to have learnt features of RBC at the individual object level. However if we check the CAM for RBC in Fig. 8 (left), we observe that the CAM generated is not aligned with the object presence. Whereas for the synthetic model, the CAM generated for RBC is correctly aligned with the object presence. This shows that synthetic data has forced the model to focus on object-level features while performing gradient descent on discriminative features. This model learning now seems more logical, which helps to instill user confidence and promote increased acceptability of deep learning solutions within the medical community, by making it more explainable.

### 4.2. Object detection algorithm

Inspired by the result of CAM using synthetic data, we started training object detection models. We hypothesized that every deep learning model can be forced to learn object level features if the same training strategy as proposed in this paper is used. We used an open-source implementation[1] of YOLO2 [13] for training with respect to object detection. As done previously, two models were trained, one with real images and another with synthetic images keeping the model parameters same. For training, we used a learning rate of 0.0005. Fig. 9 shows the results on the test set.

---
[1] https://github.com/experiencor/keras-yolo2

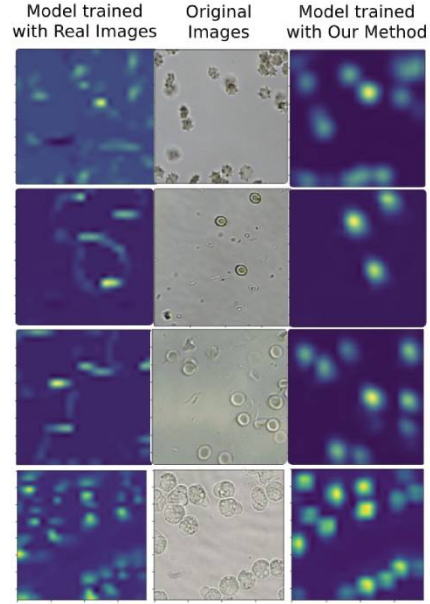

Figure 8. Class activation maps CAM [19] for weakly supervised model. (Middle) Real Test images of size 384X384 pixels. (Left) CAM for model trained with real images only. As is clear from the image, the activations are not aligned with object localization. (Right) CAM for model trained with proposed method. Here, activations are aligned with object localization proving the superiority of our method.

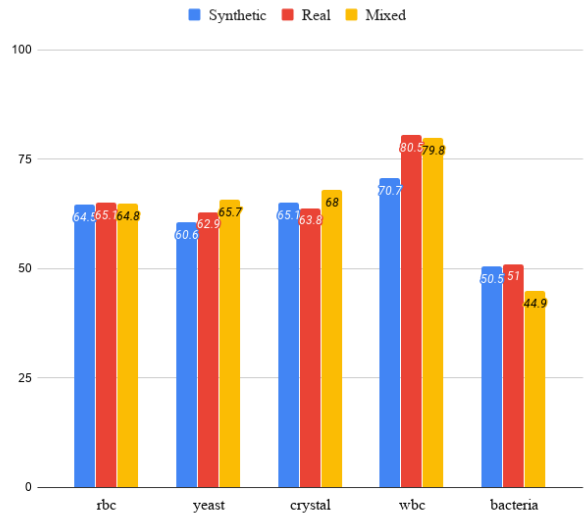

Figure 9. F1-score for object detection. Recall and precision on the test bed is calculated, followed by F1-score. Scores for bacteria are low in all methods as it is a small biological object with very less visual features.

As we can see from Fig. 9, F1-score for the model trained on synthetic images is very competitive to the model trained on real images. The advantage of synthetic images is reflected in the time saved during annotations while also overcoming the need for exhaustive annotations. The Real image dataset created for this experiment consisted of only 8k images. However annotating tens of thousands of images with bounding boxes in a given time-frame and with high accuracy is a difficult task to achieve. Our method helps to overcome such practical challenges.

We also trained a model with 10% of real images mixed with created synthetic images. As can be seen from Fig. 9, model performance has improved and surpassed for 2 classes, and equaled the result of the model trained on real images for 1 class thus establishing the superiority of our method. (F1-score below 50% ((bacteria) is not satisfactory. In our work, we found that we have sufficiently good recall for bacteria but the corresponding precision is low. This is because small objects have less visual features which result in few discriminative features also. Another point to be noted is that urine images contain many artifacts (medically insignificant objects) which look similar to bacteria (small object), and are also predicted as bacteria. So the model is able to identify a sufficiently high number of bacteria but at the same time it also predicts a significant number of false positives.)

## 5. Conclusion

We presented the use of synthetic patches for model training which helps to reduce the annotation time drastically and overcomes the need for exhaustive annotations while still producing promising results. By forcing the model to learn object-level features, this method is able to produce explainable CAM. It also helps to generate a large dataset for object detection training. Mixing synthetic data with a small percentage of real data helps to train a model which even outperforms the model trained only on real images. Future work to be explored focuses on improving object detection for small objects like bacteria, which contain less visual features. Also, since the batch statistics of the synthetic dataset are different from the real dataset, more experiments need to be conducted to establish the effect of mixing real data with synthetic data, and if changing the parameters of batch normalization as suggested by Li *et al.*[10] would further help improve the model predictions.